# A novice looks at emotional cognition

Rajendra K. Bera[1]

Acadinnet Education Services India Pvt. Ltd., Bangalore, India

Modeling emotional-cognition is in a nascent stage and therefore wide-open for new ideas and discussions. In this paper the author looks at the modeling problem by bringing in ideas from axiomatic mathematics, information theory, computer science, molecular biology, non-linear dynamical systems and quantum computing and explains how ideas from these disciplines may have applications in modeling emotional-cognition.

## 1    Introduction

As a researcher and inventor with a multidisciplinary background, when I glanced at the literature on emotional-cognition I found it was nascent, a little unsteady on the jargon, with no formal definitions of important key words (e.g., awareness, consciousness, intelligence, understanding, mind, etc.), and a lot of hand-waving kind of modeling. Not surprising. The subject is new and complex and will take decades to mature. Just the right scene for someone like me to present some speculative thoughts.

I begin with the belief that 'intelligence' requires 'understanding', which in turn requires 'awareness' and 'consciousness'. I further believe that their manifestation requires a 'mind' and the material support for the mind is the brain. We understand the terms spelled in quotes only intuitively, and our reaction to them is emotionally-cognitive in a statistical way. And finally, without some form of language the mind cannot function and therefore the mind and computations are closely linked. The brain computes, the mind interprets using language.

To guide my thoughts, I chose a few key words:

> emotion, cognition, memory, information,
> information processing, and actuators.

My speculative model has the following characteristics: Humans are non-linear chemical systems with numerous adaptive feed-back and feed-forward loops governed by laws of Nature. At the cellular level information is stored in the DNA, which information allows the cell to differentiate and replicate through resources drawn from the environment and within itself before enacting a programmed death. At the level of a phenotype (genotype

---

[1] Address for communication: Rajendra K. Bera, Chief Mentor, Acadinnet Education Services India Pvt. Ltd., B1/S1 Ganga Chelston, Silver Spring Road, Varthur Road, Munnekolala, Bangalore 560037, India.

+ environment + random variations), emotional-cognitive information is stored in and acted upon by a general-purpose neural network, some parts of which may function autonomously from time-to-time.

## 2 Abstracting out emotion and intelligence

As a first step I simplify my system by abstracting out emotion and intelligence! So I am, for the moment, essentially left with information, and information processing. There are two remarkable developments in information theory due to Claude Shannon[2] (Bell Labs, 1948) and Rolf Landauer[3] (IBM, 1961) that are crucial to this simplification. Shannon provided a mathematical definition of the concept of information, and Landauer showed that 'information is physical'.

### 2.1 Information

Shannon's genius lay in recognizing that an information source is someone or something that generates messages in a *statistical* fashion. For example, to a listener, a speaker making a statement very slowly, one letter at a time, each letter would appear as if it was an element of a random sequence although the speaker's choice may depend on what has been uttered before, while for other letters there may be a considerable amount of latitude. Shannon equated information with *uncertainty*. He then argued (as only a genius can) that suppose we have a set of possible utterances of letters whose probability of occurrences are $p_1, p_2, …, p_n$ in an ensemble of messages. That is, we *only* know the probabilities of any of the $n$ letters being uttered. Can we then find a measure of how much "choice" is there in the selection of letters or how uncertain we are of a letter being uttered?

Shannon answered with a mathematical measure: the Shannon entropy $H$ of the set of probabilities $p_1, p_2, …, p_n$. It has the form

$$H = -K \sum p_i \log p_i. \quad (K \text{ is a positive constant})$$

What is stunning about this formula is its uncanny resemblance to entropy as defined in statistical thermodynamics. Shannon had found an analogy between the random motion of molecules in a gas and randomly generated messages!

It can be shown that Shannon entropy, thermodynamic entropy, and Boltzmann's statistical definition of entropy are related to each other. Even more surprisingly, $H$ quantifies the resources needed to store and transmit information. $H$ is now the standard measure of the information contained in a message (after accounting for redundancies, statistical properties of a language, etc.). $H$ does not measure the degree of accuracy; it measures the degree of degeneracy of a system.

---

[2] Shannon, C.E., A mathematical theory of communication, The Bell System Technical Journal, Vol. 27, July, October, 1948, pp. 379–423, 623–656, http://cm.bell-labs.com/cm/ms/what/shannonday/shannon1948.pdf (reprinted with corrections).
[3] Landauer, R., Irreversibility and Heat Generation in the Computing Process, IBM Journal of Research and Development, Vol. 5, No. 3, 1961, pp. 183-191, http://www.cs.duke.edu/~reif/courses/complectures/AltModelsComp/Landauer/Landauer.irreversibility.pdf.

The next breakthrough came from Landauer who emphatically announced that 'information is physical'. For information to exist, it must be encoded in a physical system. Information theory therefore cannot be a purely mathematical concept; it must be tied to the laws of physics. Hence the laws of physics, classical or quantum, as applicable to the information carrying device, will result in different information processing capabilities.

Landauer's genius was to show that there is a fundamental asymmetry in the way Nature allows us to process information. In fact, he proved the surprising result that all but one operation required in computation could be performed in a reversible manner.[4] E.g., copying classical information can be done thermodynamically reversibly, but when information is erased there is a minimum energy cost: $kT \ln 2$ per classical bit to be paid (here $k$ is the Boltzmann constant, and $T$ is the temperature of the environment in Kelvin). That is, the erasure of information is inevitably accompanied by the generation of heat. This remarkable result is the equivalent of the second law of thermodynamics. Indeed, Landauer's principle related to the erasure of information provides a bridge between information theory and physics.

Landauer's insight that 'information is physical' implied that information and free energy are interconvertible. Indeed, computers may be viewed as engines for transforming free energy into waste heat and mathematical work.[5] An experimental demonstration of information-to-energy conversion was reported by S. Toyabe, et al[6] in *Nature Physics*, and by A. Bérut, et al[7] in *Nature*. So the conversion is real.

I shall return to information and energy in Section 4 to show a possible link between information and emotion.

## 2.2    Information processing

There are two landmark developments in information processing: limitations of formal axiomatic systems and limitations of universal Turing machines.

The structure of an axiomatic system is that it begins with a specification (a set of symbols and a grammar or typographical rules for combining the symbols into statements) regarding the construction of syntactically correct statements (well-formed formulas), a set of axioms (propositions regarded as self-evident truths; postulates; they may be any number including zero and infinity), and a finite set of inference rules that allow theorems to be generated using the axioms and previously derived theorems. As far as I know, creating an axiomatic system is a non-mathematical and a highly intelligent act. Developing a sequence of theorems with a specific goal in mind (developing algorithms) is also a highly intelligent act. However, executing an algorithm, once developed, is mechanizable and does not require intelligence, in fact, none at all.

---

[4] *See*, *e.g.*, Feynman, R.P., The Feynman Lectures on Computation, Westview, 1999, http://www.scribd.com/doc/52657907/Feynman-Lectures-on-Computation.

[5] *See* Bennett, C.H., The Thermodynamics of Computation—A Review, International Journal of Theoretical Physics, Vol. 21, No. 12, 1982, pp. 905-940, http://www.cc.gatech.edu/computing/nano/documents/Bennett%20-%20The%20Thermodynamics%20Of%20Computation.pdf; and
Bennett, C.H., Demons, Engines and the Second Law, *Scientific American*, Vol. 257, No. 5, pp. 108-116, November 1987, http://physics.kenyon.edu/people/schumacher/Physics110/DemonsEnginesAndSecondLaw87.pdf.

[6] Toyabe, S., Sagawa, T., Ueda, T., Muneyuki, E., and Sano, M., Experimental demonstration of information-to-energy conversion and validation of the generalized Jarzynski equality, Nature Physics, Vol. 6, pp. 988–992, 14 November 2010 (online).
For an earlier version see http://arxiv.org/pdf/1009.5287v2.pdf.

[7] Bérut, A., Arakelyan, A., Petrosyan, A., Ciliberto, S., Dillenschneider, R., and Lutz, E., Experimental verification of Landauer's principle linking information and thermodynamics, Nature, Vol. 483, 08 March 2012, pp. 187-189, http://211.144.68.84:9998/91keshi/Public/File/34/483-7388/pdf/nature10872.pdf.

In 1930, Kurt Gödel[8], in what many believe to be the most important result in mathematics, showed that no formal axiomatic system of mathematics sufficiently strong to allow one to do basic arithmetic (such as Peano arithmetic) can be at once consistent (i.e., it will never produce contradictory results) and complete (i.e., the system should be capable of testing the truth or falsity of all possible propositions in mathematics). In essence, the mathematics we use has a deep flaw, yet surprisingly it has not prevented us from using mathematics very effectively in science or commerce. As a matter of great importance, note that any axiomatic system can be converted into an arithmetical system using an appropriate mapping, and so we step into the realm of computing.

While Gödel showed that there are limits to what we can, in principle, prove, Alan Turing[9] in 1936 showed that there are limits to what we can compute. He did this by defining an abstract model of a human engaged in doing arithmetic without the benefit of insight. The task of computing, he felt, was nothing but calculations performed by a human mathematician who has unlimited time and energy, an unlimited supply of paper and pencils, perfect concentration, and who worked according to some algorithmic or 'rule-of-thumb' method. This abstract machine, now called the universal Turing machine (UTM), is simple enough to allow mathematicians to prove theorems about its computational capabilities and yet sufficiently complex to accommodate any actual classical digital computer, no matter how it is implemented (supercomputers included), with the idealization that the computer must have access to an unlimited storage capacity.

Turing, with a touch of genius, then showed that there exist uncomputable problems, the famous 'halting problem' in computer science being one of them. (Calculating random numbers is another example.) Turing, in essence, established an isomorphism between formal and computational systems, as shown in Table 1. From uncomputability, he was then able to deduce incompleteness (in the sense of Gödel) and extend Gödel's incompleteness theorem. In essence, Turing showed that undecidability in mathematics was even more widespread than had been anticipated.

---

[8] Gödel, K., a young Austrian mathematician (at age 24), first announced the result to the Vienna Academy of Sciences in 1930 and later published it in a paper, Kurt Gödel, Über formal unentseheidbare Sätze der Principia Mathematica und verwandter Systeme I, Monatshefte für Mathematik und Physik, Vol. 38, pp. 173-198 (Leipzig: 1931). (On formally undecidable propositions of Principia Mathematica and related systems I.) Part II of the paper was never published.
Visit http://www.w-k-essler.de/pdfs/goedel.pdf for the original paper in German.
Visit http://jacqkrol.x10.mx/assets/articles/godel-1931.pdf for an English translation by B. Meltzer.

[9] Turing, A., On computable numbers, with an application to the Entscheidungsproblem, Proceedings of the London Mathematical Society, Series 2, Vol. 42, 1936, pp. 230-265 (Errata: *Proceedings of the London Mathematical Society*, Series 2, Vol. 43, 1937, pp. 544-546.). Reproduced in M. D. Davis, The Undecidable, Raven Press, Hewlett, New York, 1965. Available at http://www.abelard.org/turpap2/tp2-ie.asp.

| Table 1 Isomorphism between formal axiomatic system and computational system. ||
|---|---|
| **Formal system** | **Computational system** |
| Axioms | Program input or initial state |
| Rules of inference | Program interpreter |
| Theorem(s) | Program output |
| Derivation | Computation |

## 3      Bringing in intelligence; isomorphism

We can now bring in intelligence into the picture using the notion of isomorphism. We say two systems are isomorphic if they can be mapped onto each other such that for each part of one system there is a corresponding part in the other system. Note that isomorphism is an information preserving mapping and not a 'meaning' preserving mapping. To detect isomorphism requires intelligence, usually of a very high order.

> The perception of an isomorphism between two known structures is a significant advance in knowledge – and I claim that it is such perceptions of isomorphisms which create *meanings* in the minds of people.
> 
> — Douglas Hofstadter, Gödel, Escher, Bach (1979)

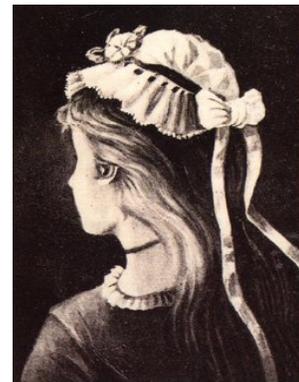

Here is an unusual example of isomorphism. What do you see in the picture?[10] A pretty girl or an old woman? For each point describing the pretty girl, there is a corresponding point describing the old woman. The two are isomorphic! This may give you some idea as to what quantum physicists mean when they say that an electron can be simultaneously in a spin-up and spin-down state. (By the way, what was your emotional-cognitive reaction to what you saw in the picture? Same information, same reaction?)

Since isomorphism may come in a variety of forms, the crucial question is: "Given an unknown formal system, how do you discover some hidden meaning in it?" In short, how do we assign interpretations or meanings to the symbols of a formal axiomatic system (or to numbers in a computational system) in a meaningful way – i.e., in such a way that a higher-level correspondence emerges between true statements about the world and abstract theorems? The discovery seems to depend on *intelligent* guesswork! When you hit upon a good or a meaningful choice everything suddenly seems to fall in place, everything seems to feel right, the puzzle is solved. *That is the moment of excitement*, the *aha*! moment.[11]

---

[10] Source of picture: My wife and my mother-in-law,
http://en.wikipedia.org/wiki/File:German_postcard_from_1888.png.

Reality and formal systems are independent. No one needs to be aware that there is an isomorphism between the two. Note that much of the mathematics used by physicists was not created for use by physicists! Physicists found it lying around when they needed it. Indeed, Eugene P. Wigner wrote a famous paper titled "*The unreasonable effectiveness of mathematics in the natural sciences*."[12]

I firmly believe that meaning is always manufactured by the interaction of a mind with a message. Sometimes we may strongly believe there is an isomorphism but may not be quite capable of discovering the mapping.

Here is an example. The information in the DNA—a genotype—is converted into a physical organism—a phenotype—guided by a set of enormously complex cycles of chemical reactions and feedback loops. The emerging phenotype has no apparent similarity between its physical characteristics and its genotype. Yet molecular biologists attribute the physical structure of the organism to the information in the DNA, and to that alone. In essence, they claim that the information carried by the DNA is isomorphic to the phenotype's structure.

However, this isomorphism is highly exotic. We do not yet know how to divide the phenotype and genotype into parts which can be mapped onto each other, unlike as, say, in the case of a music disk and the notes produced by its player. Of course, attempting to map musical notes to the emotions they raise in us is, once again, highly exotic. The difficulty lies in the fact that "genetic meaning"—that is, information about phenotype structure—is spread all through the small parts (e.g., genes) of a molecule of DNA, although nobody understands the language yet.

Molecular biologists and mathematicians are becoming increasingly sure that "life is a partnership between genes and mathematics". But the nature of that partnership remains largely unknown. Nevertheless, it is a delight that cells are so elegantly designed that for many purposes one can totally ignore their chemistry and think just about their logic — "the molecular logic of the living state"[13] as Albert Lehninger put it — that you do not have to be a biochemist to understand it.

If there is mathematical logic in living things then one naturally seeks to determine what formal system governs life. In other words, what is the mathematical basis around which information in the DNA is stored and how is that information used by the rest of the cell's machinery to enable the phenotype to do the myriad of things that it does, including emotional-cognition.

---

[11] Hofstadter, D.R., Gödel, Escher, Bach: An Eternal Golden Braid, Basic Books, 1979, p. 50.

[12] Wigner, E., The Unreasonable Effectiveness of Mathematics in the Natural Sciences, Communications in Pure and Applied Mathematics, Vol. 13, No. 1 (February 1960), http://www.physik.uni-wuerzburg.de/fileadmin/tp3/QM/wigner.pdf.

[13] Lehninger, A.L., Biochemistry, 2nd edn., Worth Publishers, New York, 1975. The phrase appears in "living organisms are composed of lifeless molecules … that conform to all the laws of chemistry but interact with each other in accordance with another set of principles—the molecular logic of the living state." Quote taken from I.D. Campbell, The Croonian lecture 2006: Structure of the living cell, Phil. Trans. R. Soc. B (2008) **363**, 2379-2391, http://rstb.royalsocietypublishing.org/content/363/1502/2379.full.pdf.

## 4     Information and energy

We now return to information and energy. We know that polarized photons can be encoded to carry binary information, say, in horizontal and vertical polarizations. This leads to the tantalizing possibility of a link between axiomatic systems and the ability of the brain-mind system to interpret and therefore contextually extract meaning from formal systems. That link appears to be energy and its ability to change forms (e.g., from mass to energy and vice-versa according to Einstein's formula, $E = mc^2$, a fact used in the design of atom and H-bombs for converting mass-to-energy, and in particle accelerators for converting energy-to-mass). The photon is quantized energy and is the carrier of the electromagnetic force; it is a quantum mechanical object. So it appears that the answer to understanding emotional-cognitive phenomena may finally lie in quantum computing, provided we can show how emotion and energy are related.

It may be possible to determine this relationship one day since we all experience the ebb and flow of energy correlated with our changing emotions. And our individual emotions are correlated in the way we interpret the information received by us. Since a given message may be interpreted differently by different people, individual interpretations in a group of people will display a statistical distribution. A quantum mechanical analog of this is that a message can be in a superposed state of interpretations till it is interpreted (measured) by individuals. I believe interpretation is governed by a probabilistic postulate, akin to the probabilistic measurement postulate in quantum mechanics (see Section 6.3). I use the word 'postulate' deliberately because I do not think we are likely to discover the mechanism by which an individual chooses to express himself specifically from a possible number of emotions that a message may evoke.

## 5     Emotion and cognition

Our understanding of the role of emotion in human behavior is primitive. Intellect and emotions are inseparably controlled by our individual and collective (group of individuals) neural systems. Moreover, our emotional and intellectual activities are often results of group efforts or akin to group efforts, and those results sometimes produce statistical patterns (based on the practice of collecting and analyzing numerical data in large quantities).

I believe that while the emotional-cognition problem has an underlying layer of determinism, which one day physicists might discover, at a human observational level the issue must be understood in terms of a probability distribution of possible outcomes. The probability distribution comes from the emotional element that we bring to any human cognition phenomenon.

When did emotion and cognition begin to rule our lives? My belief is that with an elementary neural system with some hard-wired capacity for information processing for instinctive reaction and an innate ability to correlate events and information, especially repetitive phenomena, emotion and cognition would have developed.

For thinking, language had to develop, and for rational thinking logic had to develop as a special case of emotionless cognition. However, to build the foundations of rational thought emotion was required to enunciate the unprovable beliefs (axioms) that could be shared by all humanity. Emotions help us select beliefs with some intrinsic beauty in them, usually in some form of symmetry, and this intrinsic beauty appears to be a shared quality among humans when seeking isomorphisms. I believe that generation and progression of ideas cannot occur without emotional involvement, and that this involvement is governed by, as yet, undiscovered statistical laws.

The path I might choose to follow in defining emotion and cognition is to see if I can draw analogies from concepts drawn from quantum mechanics, non-linear dynamical systems, and molecular biology. Of course, I am not the first one to do so. The far more illustrious Roger Penrose[14] had attempted to do so well before me. (I disagree with some of his views but that is to be expected in a nascent subject.)

My starting point is that conscious awareness—of pain, happiness, love, hatred, aesthetic sensibility, will, understanding, etc.—is a feature of the brain's physical action; and "whereas any physical action can be simulated computationally, computational simulation cannot by itself evoke awareness"[15]. I hold this view since consciousness can be switched on and off using chemicals, e.g., anesthesia, without loss of memory. Even moods can be changed by chemicals. In short, as 'information is physical', so is conscious awareness physical. I regard it as some kind of 'emergent property' resulting from a computation made by a specially constructed physical object, e.g., the human body, and its neural network, in particular. To understand this 'emergent property' I wish to draw upon concepts from quantum mechanics, and non-linear dynamical systems (chaos and strange attractors).

## 6 Chaos, strange attractor, quantum mechanics

### 6.1 Chaos

Chaos manifests itself in the form of a wild and unpredictable dynamics. Chaotic systems can be mathematically modeled and when so modeled they are completely deterministic and computational (or algorithmic), yet their behavior when exhibiting chaos *appears* completely unpredictable. That is because a system in chaos is in an unstable state and computing its trajectory becomes extremely sensitive to the accuracy with which the system's initial condition is prescribed (e.g., rounding errors in providing input data or rounding errors made during computations). It is obviously impossible to provide initial conditions with infinite accuracy. This is the famous 'butterfly effect' in chaos theory.

> Chaos: When the present determines the future, but the approximate present does not approximately determine the future. — Edward Lorenz (2005)[16]

While it is not practicable to predict computationally the actual outcome of a chaotic system, a mathematical simulation of a *typical* outcome is perfectly achievable. An example of a chaotic system is the weather. Given even reasonably good input data to a weather model, the predicted weather may well not be the weather that actually occurs, but it will still be perfectly plausible as *a* weather. (The seasons in any given year are similar to those of any other year but may vary widely in hourly detail.) This should not worry us in modeling emotional-cognitive behavior since our goal is not to simulate the behavior of any particular individual; we would be quite happy with the simulation of just *an* individual.

---

[14] Penrose, R., Shadows of the Mind, Oxford University Press, 1994.
[15] Penrose, R., Shadows of the Mind, Vintage, 1995, Chapter 1, p. 12. Penrose mentions this as a possible viewpoint, but he himself believes that "Appropriate physical action of the brain evokes awareness, but this physical action cannot even be properly simulated computationally."
[16] As quoted by Danforth, C.M., Chaos in an Atmosphere Hanging on a Wall, Mathematics of Planet Earth 2013, April 2013, http://mpe2013.org/2013/03/17/chaos-in-an-atmosphere-hanging-on-a-wall/.

## 6.2     Strange attractors

In mathematics attractors are states of a dynamical system to which the system is attracted to over time. When a system reaches an attractor it remains in its vicinity if slightly disturbed and eventually returns to it given sufficient undisturbed time. Known attractors are: fixed point, limit cycle, torus, and strange attractor. A system is said to settle onto a strange attractor – a set of states on which it wanders forever, never stopping or repeating. Such erratic, aperiodic motion is considered chaotic if two nearby states flow away from each other exponentially fast. Long-term prediction is impossible in a real chaotic system (e.g., the weather) because of this exponential amplification of small uncertainties or measurement errors. Instability of motion associated with chaos allows the system to explore continuously in state space, thereby creating information and complexity in the form of aperiodic sequences of states.

## 6.3     Quantum mechanics

A quantum object can simultaneously be in more than one state if unobserved or not measured, but when observed it randomly collapses to one of its sub-states[17] (remember the picture 'My wife and my mother-in-law'). The collapse mechanism is unknown and the moment of collapse is indeterminate. Two quantum systems (even as simple as two one-photon systems) can get *entangled* in a manner that independent of the distance between them (even if light years apart), a change in one system will instantly bring about a predictable change in the other. (This allows quantum states to be teleported, say, from one photon to another.) Finally, there are pairs of complementary variables (e.g., position and momentum), say, of an electron, which cannot be measured completely accurately (even with perfect instruments) simultaneously (Heisenberg's uncertainty principle).

None of these – superposition, collapse, entanglement, and the uncertainty principle – appear in classical physics. We still do not know completely how the laws of quantum mechanics transition to the laws of classical physics. The answer almost certainly lies in the mechanism that leads to the collapse of a quantum system when measured. This mechanism is quite likely to be a subtle non-computational (but still mathematical) physical scheme.[18] Application of quantum mechanics to biology (the subject is called quantum biology) is at a nascent stage with few concrete results. For example, we know that pigeons have "an identifiable neural substrate for processing magnetic sense information"[19]. In the coming decades I hope to see the subject provide some deep insights about the functioning of the human neural system.

---

[17] This random collapse of a quantum system is a highly unusual feature of Nature. It is the origin of all the probabilities one talks about in quantum mechanics. The evolution of a quantum system according to the Schrödinger equation is completely deterministic. It is only when a classical measurement on the system is made that probabilities kick in. *See* Born, M., The statistical interpretation of quantum mechanics, Nobel Lecture, December 11, 1954. http://nobelprize.org/nobel_prizes/physics/laureates/1954/born-lecture.pdf.

## 7    The long road ahead

Human neural networks (essentially brains) are capable of commanding highly non-linear adaptive dynamics and forming concepts. I believe concepts appear as strange attractors in neural network dynamics. Recall is the ability of the brain to retrace a neural path traversed at least once in the past. Insight is the ability to chain concepts. Physical actuators are muscles.

The human body receives inputs from the external world via the senses, and from the neural network via memory recall, and chaining of concepts. The human neural network is capable of dealing with both deductive and inductive logic and of performing, at least, as a probabilistic Turing machine (PTM). (PTM is a UTM in which some transitions are based on random choices from among finitely many alternatives. PTM can be simulated by a UTM.) I tentatively believe neural connections can be understood properly via quantum chemistry while concept formation can be understood using classical non-linear dynamics.

Much of our understanding of the brain is rudimentary. It is essentially the model put forth by Warren McCullouch (a neuroscientist) and Walter Pitts (a logician) in 1943 that became the foundation for future artificial neural network (ANN) research.[20] In their highly simplified model neurons and their connecting synapses seem to play a role essentially similar to those of transistors and wires of those days. My belief is that to understand the brain we will need to understand a far more complex model that includes quantum systems that retain their quantum nature (quantum parallelism, non-locality, counterfactuality[21]) at a much larger spatial and temporal scale than is evident today and the circumstances under which they can do so. Also, how these systems collapse to a classical state.

Emotions convey information and information is physical. The physical manifestation of emotion is a correlated dynamic geometrical pattern a living system displays on its person in response to certain information interpretations produced by the living system. Cognition is a purely neural network activity. Information storage and information processing are purely chemical activities obeying the laws of quantum chemistry.

---

[18] *See*, *e.g.*, Bera, R.K., and Menon, V., A new interpretation of superposition, entanglement, and measurement in quantum mechanics, arXiv:0908.0957v1 [quant-ph], 07 August 2009, http://arxiv.org/abs/0908.0957.

[19] Lambert, N., Chen, Y-N, Cheng, Y-C. Li, C-M, Chen, G-Y, and Nori, F., Quantum biology, Nature Physics, Vol. 9, January 2013, pp. 10-18, http://www.readcube.com/articles/10.1038/nphys2474.

[20] McCullouch, W.S, and Pitts, W.H., A logical calculus of the ideas immanent in nervous activity, Bulletin of Mathematical Biophysics 5:115-133, http://www.cse.chalmers.se/~coquand/AUTOMATA/mcp.pdf.

[21] Counterfactuals are things that might have happened, although they did not in fact happen. *See*, *e.g.*, Vaidman, L., Counter-factuals in quantum mechanics, arXiv:0709.03401, 04 September 2007, http://arxiv.org/pdf/0709.0340v1.pdf.

Information processing capabilities of a human neural network are at least as powerful as a PTM.

An almost pure emotional system is a new-born baby, which from birth begins to evolve to an emotionally-cognitive system. The sophisticated end of cognition deals with abstract processing (axiomatic reasoning). The famous Indian mathematician Srinivasa Ramanujan is an outstanding example.

For humans, emotion and cognition are deeply entwined and it surely has something to do with the way the human brain is organized. I agree with Luiz Pessoa[22] "that complex cognitive–emotional behaviours have their basis in dynamic coalitions of networks of brain areas, none of which should be conceptualized as specifically affective or cognitive. Central to cognitive–emotional interactions are brain areas with a high degree of connectivity, called hubs, which are critical for regulating the flow and integration of information between regions."

When it comes to emotions, our current knowledge forces us to use heuristic plausibility arguments. It may well be that modeling emotions with certainty is an unattainable limit point as understood in mathematics. It may also be that reading the Mahabharata[23], the plays of Shakespeare, the works of Homer, Hofstadter's 'Gödel, Escher, Bach', novels by Charles Dickens and Jane Austen will educate us more in understanding emotional-cognition than the current research literature on the subject. I say this with all humility. Only Isaac Newton could deduce the law of gravitation and proclaim that the Earth and the Moon were literally falling into each other by seeing an apple fall. Similar insights about emotional-cognition have so far eluded me.

---

[22] Pessoa, L., On the relationship between emotion and cognition, Nature Reviews Neuroscience, Vol. 9, February 2008, pp. 148-158, http://www.psych.nmsu.edu/~jkroger/lab/COURSES/375/cognition/2008%20Pessoa%20-%20On%20the%20relationship%20between%20emotion%20and%20cognition.pdf.

[23] An epic of ancient India. It is roughly ten times the length of the *Iliad* and *Odyssey* combined.